\documentclass{acm_proc_article-sp}
\usepackage{mathtools}
\DeclarePairedDelimiter\abs{\lvert}{\rvert}

\begin{document}

\title{A review of various short-term traffic speed forecasting models}

\numberofauthors{2}
\author{
\alignauthor John Boaz Lee\\
       \affaddr{Information Sys. and Computer Science Dept.}\\
       \affaddr{Ateneo de Manila University, Philippines}\\
       \email{jlee@ateneo.edu}
\alignauthor Kardi Teknomo\\
       \affaddr{Information Sys. and Computer Science Dept.}\\
       \affaddr{Ateneo de Manila University, Philippines}\\
       \email{teknomo@gmail.com}
}

\maketitle
\begin{abstract}
The widespread adoption of smartphones in recent years has made it possible for us to collect large amounts of traffic data. Special software installed on the phones of drivers allow us to gather GPS trajectories of their vehicles on the road network. In this paper, we simulate the trajectories of multiple agents on a road network and use various models to forecast the short-term traffic speed of various links. Our results show that traditional techniques like multiple regression and artificial neural networks work well but simpler adaptive models that do not require prior training also perform comparatively well. 
\end{abstract}

\section{Introduction}
In the past, intelligent transportation systems (ITS) relied on speed sensors installed in various locations to monitor traffic conditions. Because of the relatively large number of roads in urban areas and the high cost to set-up and maintain these traditional speed sensors, most ITS operators are only able to monitor the traffic conditions of urban areas partially. Now that the use of GPS-enabled devices has become quite pervasive, vehicles with these devices can be treated as a new kind of mobile sensor. In the city of Beijing alone, more than 50,000 licensed taxicabs (many of which are GPS-equipped) generate over 1 million vehicle trajectories per day \cite{taxi}. This large amount of data, when processed and analysed, can be used in various applications.\\\\
Vehicle trajectories have been used to solve various problems in the domain of urban computing, which is an emerging field of study concerned with the application of technology in the urban setting \cite{urban}. In \cite{taxi}, for instance, researchers used the trajectories generated by around 30,000 taxicabs to identify defects in urban planning and regions with salient traffic problems. The same type of data was also used to identify unusual traffic patterns in another study \cite{anomaly}. This allowed researchers to automatically identify special events, like celebrations, protests, road closures, or large-scale business promotions, that disrupted the normal flow of traffic in an area.\\\\
Wei et al. \cite{pop_routes} devised an algorithm that could analyse vehicle trajectories to identify popular routes. Yuan et al. \cite{tdrive}, on the other hand, analysed the trajectories of taxi drivers to give driving directions. By leveraging the intelligence of taxi drivers, the developed system \cite{tdrive} was able to suggest routes that were often faster than routes suggested by other approaches.\\\\
Now that real-time traffic information has become more accessible due to the relative ease in collecting GPS trajectories of vehicles, the next logical step is to forecast future traffic condition \cite{pred1}. This is a problem that has many practical applications. For instance, a system that suggests routes to drivers can take advantage of traffic forecasts to recommend routes that are predicted to be less congested in the near future.\\\\
The problem of forecasting short-term traffic condition is one that is well-studied in the literature; many different approaches have already been proposed. In some of the earlier studies, univariate time series models like the Box-Jenkins autoregressive integrated moving average (ARIMA) and various exponential smoothing models were used to forecast short-term traffic \cite{svr, arima, es}. More recently, various approaches utilizing neural networks were introduced \cite{neural1, neural2, neural3}. An approach based on dynamic wavelet neural networks was proposed in \cite{neural3} while one that combined fuzzy logic, wavelet transforms, and neural networks was studied in a paper by Xiao et al. \cite{neural4}.\\\\
More advanced approaches include the model proposed by Min and Wynter \cite{pred1} that utilizes a spatial-temporal autoregressive model to predict short-term traffic. Castro-Neto et al. \cite{svr}, on the other hand, presents a supervised statistical learning technique called the Online Support Vector machine for Regression that is able to predict traffic flow under both typical and atypical conditions. Finally, \cite{hybrid} introduces a model that combines clustering and regression-based techniques to predict traffic speed.\\\\
In this paper, we survey various models for forecasting short-term traffic speed. The goal is not to devise a novel approach for solving the problem of traffic speed forecasting. Instead, we test several techniques to discover which works best with GPS trajectory data in practice.\\\\
The rest of the paper is structured as follows. In the next section we talk about the data and the models to be tested. We then explain the experiments and elaborate on the experimental results in the third section. Finally, we conclude the paper with some suggestions for future work.

\begin{figure}[t]
	\centering
	\includegraphics[width=0.9\columnwidth]{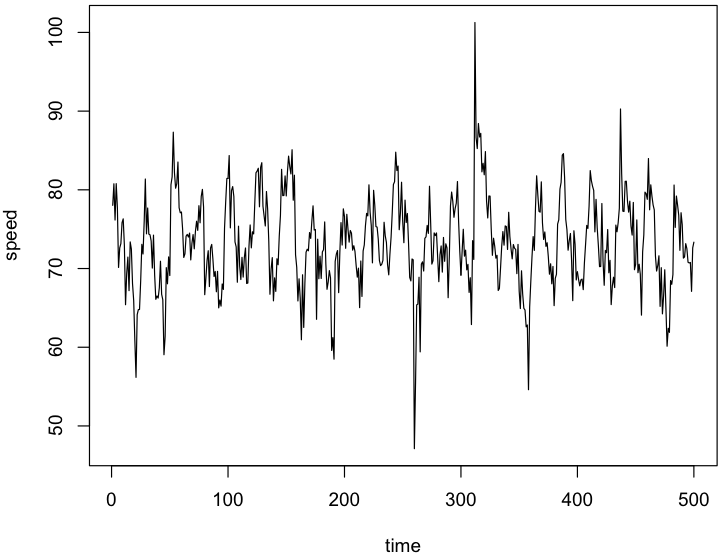}
	\caption{Simulated speed time series generated using a fitted SARIMA model.}
\end{figure}

\section{Methodology}
\subsection{Dataset}
Although the researchers have already implemented a system to collect GPS trajectories of vehicles in Metro Manila, the volume of data collected at the time of the study was insufficient for forecasting. Because of this, a synthetic vehicle trajectory dataset was constructed.\\\\
A graph to represent a road network was first created, time series speed data was then generated for each link or road in the network. Since real-world traffic speed follow cyclical and seasonal patterns, the generated data should also exhibit these features. To ensure this, we fitted a seasonal ARIMA model \cite{fpp} on real-world traffic speed data taken from the North Luzon Expressway \cite{nlex}. This model was then used to create the simulated speed data. An example of a simulated speed time series is shown in fig. 1.\\\\
We then simulated the trajectories of multiple agents on the graph. Even though the movements of the agents across the network were random, they were still subject to several constraints. First of all, the number of trajectories utilizing a certain link on the network is inversely proportional to the underlying traffic speed of the link at the current time point. Intuitively, the reason for this is because we can expect to capture the trajectories of more vehicles along a heavily congested road than a road that has light traffic and high speed. Secondly, the actual speed logged by a vehicle's trajectory as it traverses a link is drawn from a normal speed distribution with mean equals to the speed of the link at the current time point and standard deviation of 10. We can expect the speed of different trajectories traversing the same link to be slightly different just as different drivers drive their vehicles at different preferred speeds. Moreover, as shown in various studies \cite{dist1}, traffic speed follow a normal distribution.
\subsection{Supervised learning}
The first set of models we chose to test were supervised learning models. In particular, we tested the data on neural network and multiple regression models. The input consisted of a few variables, namely the observed speed in the last three time points, and variables denoting the hour and day of the time point whose speed was being forecasted.\\\\
A model was trained for each link in the road network using data gathered from the first 6 months. In practice, to avoid the problem of concept drift \cite{drift}, the models will have to be retrained periodically. Given the fact that there are more than 100,000 links in the Metro Manila road network, retraining will have to be done by batch and during the system's off-peak hours.
\begin{table*}[t]
\begin{center}
{\small
\begin{tabular}{|l|l|l|l|}
\hline
 &\textbf{No seasonality}&\textbf{Additive seasonality}&\textbf{Multiplicative Seasonality}\\
\hline
\hline
\textbf{No Trend}&
{$\begin{aligned}[t]
            \mu_{t+h} = a_t\\
            a_t = \alpha x_t + (1 - \alpha)a_{t-1}
    \end{aligned}
$}
&
{$\begin{aligned}[t]
            \mu_{t+h} = a_t + c_{t-s+h}\\
            a_t = \alpha (x_t - c_{t-s}) + (1 - \alpha) a_{t-1}\\
	   c_t = \gamma (x_t - a_t) + (1 - \gamma) c_{t-s}
    \end{aligned}
$}
&
{$\begin{aligned}[t]
            \mu_{t+h} = a_t c_{t-s+h}\\
	   a_t = \alpha \frac{x_t}{c_{t-s}} + (1 - \alpha) a_{t-1}\\
	   c_t = \gamma \frac{x_t}{a_t} + (1 - \gamma) c_{t-s}
    \end{aligned}
$}\\
\hline
\textbf{Add. Trend}&
{$\begin{aligned}[t]
            \mu_{t+h} = a_t + b_t h\\
            a_t = \alpha x_t + (1 - \alpha)(a_{t-1} + b_{t-1})\\
	   b_t = \beta (a_t - a_{t-1}) + (1 - \beta) b_{t-1}
    \end{aligned}
$}
&
{$\begin{aligned}[t]
            \mu_{t+h} = a_t + b_t h + c_{t-s+h}\\
            a_t = \alpha (x_t - c_{t-s}) + (1 - \alpha) (a_{t-1} + b_{t-1})\\
	   b_t = \beta (a_t - a_{t-1}) + (1 - \beta) b_{t-1}\\
	   c_t = \gamma (x_t - a_t) + (1 - \gamma) c_{t-s}
    \end{aligned}
$}
&
{$\begin{aligned}[t]
            \mu_{t+h} = (a_t + b_t h) c_{t-s+h}\\
	   a_t = \alpha \frac{x_t}{c_{t-s}} + (1 - \alpha) (a_{t-1} + b_{t-1})\\
	   b_t = \beta (a_t - a_{t-1}) + (1 - \beta) b_{t-1}\\
	   c_t = \gamma \frac{x_t}{a_t} + (1 - \gamma) c_{t-s}
    \end{aligned}
$}\\
\hline
\textbf{Mult. Trend}&
{$\begin{aligned}[t]
            \mu_{t+h} = a_t b_t^h\\
            a_t = \alpha x_t + (1 - \alpha)(a_{t-1} b_{t-1})\\
	   b_t = \beta \left(\frac{a_t}{a_{t-1}}\right) + (1 - \beta) b_{t-1}
    \end{aligned}
$}
&
{$\begin{aligned}[t]
            \mu_{t+h} = a_t b_t^h + c_{t-s+h}\\
            a_t = \alpha (x_t - c_{t-s}) + (1 - \alpha) (a_{t-1} b_{t-1})\\
	   b_t = \beta \left(\frac{a_t}{a_{t-1}}\right) + (1 - \beta) b_{t-1}\\
	   c_t = \gamma (x_t - a_t) + (1 - \gamma) c_{t-s}
    \end{aligned}
$}
&
{$\begin{aligned}[t]
            \mu_{t+h} = a_t b_t^h c_{t-s+h}\\
	   a_t = \alpha \frac{x_t}{c_{t-s}} + (1 - \alpha) (a_{t-1} b_{t-1})\\
	   b_t = \beta \left(\frac{a_t}{a_{t-1}}\right) + (1 - \beta) b_{t-1}\\
	   c_t = \gamma \frac{x_t}{a_t} + (1 - \gamma) c_{t-s}
    \end{aligned}
$}\\
\hline
\end{tabular}}
\vspace{3pt}
\caption{Formula for the different exponential smoothing models.}
\label{tb:tablename}
\end{center}
\end{table*}
\subsection{Exponential smoothing}
Nine different variants of exponential smoothing were tested too. Some of the models captured seasonality and trend while others did not. The advantage of using relatively simple models like exponential smoothing is that these do not have to be trained and can be easily modified to make them adaptive or dynamic.\\\\
In its simplest form, the exponential smoothing formula can be written as 
$$\mu_t = (1-\alpha)\mu_{t-1} + \alpha x_t$$
where $x_t$ is the observed speed at time $t$, $\mu_{t-1}$ is the previous forecast and $\mu_t$ is the current forecast. The smaller the value of $\alpha$, the more dependent the forecast is on past information rather than the current data.\\\\
The above approach can be modified to include an adaptive parameter which changes over time. Instead of using a constant parameter $\alpha$, we replace it with the dynamic parameter $\alpha_t$. In this formulation, a second parameter $\beta$ will have to be introduced as well. The new formula is
\begin{align*}
\mu_t = (1-\alpha_t)\mu_{t-1} + \alpha_t x_t\\
\alpha_t =  \abs*{\frac{a_t}{b_t}}\\
a_t = \beta e_t + (1- \beta)a_{t-1}\\
b_t = \beta \abs{e_t} + (1 - \beta)b_{t-1}\\
e_t = x_t - \mu_t
\end{align*}
where $\beta$ is a parameter to control the fluctuation of the dynamic parameter $\alpha$. Here, to initialize, one can set $\mu_2 = x_1, \alpha_2 = \alpha_3 = \alpha_4 = \beta$ and $a_1 = b_1 = 0$.\\\\
When the data has trend, one can extend the original formula to a linear exponential smoothing model. Here is the model proposed by Holt \cite{fpp}
\begin{align*}
\mu_{t+h} = a_t + b_t h\\
a_t = \alpha x_t + (1 - \alpha)(a_{t-1} + b_{t-1})\\
b_t = \beta(a_t - a_{t-1}) + (1 - \beta)b_{t-1}
\end{align*}
where $h$ is the period ahead to forecast and $a_t$ is the level of the series while $b_t$ is the slope of the series at time $t$. Here, initialization is done as follows: $a_1 = x_1$ and $b_1 = x_2 - x_1$ or $b_1 = \frac{1}{3}(x_4 - x_1)$.\\\\
Finally, to capture seasonality in the data, one can use Holt and Winter's formulation \cite{fpp}
\begin{align*}
\mu_{t+h} = (a_t + b_t h)c_{t-s+h}\\
a_t = \alpha \frac{x_t}{c_{t-s}} + (1 - \alpha)(a_{t-1} + b_{t-1})\\
b_t = \beta (a_t - a_{t-1}) + (1 - \beta)b_{t-1}\\
c_t = \gamma \frac{x_t}{a_t} + (1 - \gamma)c_{t-s}
\end{align*}
here $s$ is the length of the season. In this model, the level is initialized by taking the average of the first season $a_s = \frac{1}{s} \sum_{i=1}^s x_i$, trend is initialized $b_s = \frac{1}{s} \sum_{i=1}^s \frac{x_{s+i} - x_i}{s}$, and the seasonal indices $c_i$ are initialized $c_i = \frac{x_i}{a_s}$. Table 1 shows the formulation of the different exponential smoothing methods that we will be testing. Aside from the methods listed in table 1, we also tested the adaptive version of the traditional exponential smoothing model. One can refer to \cite{fpp} for a thorough explanation of exponential smoothing.
\section{Experimental setup, results, and discussion}
We tried to forecast short-term traffic speed using the methods discussed in the previous section. Feed-forward neural networks\footnote{The implementation found in the caret package of R was utilized. Every time the model is retrained, it performs slightly differently.} were utilized and trained using the simulated data. Various versions with different decay rates and number of hidden units were tested to identify the best setup. To test the exponential smoothing models, we tested each model with different combinations of values for each parameter. The parameters, $\alpha, \beta, $ and $\gamma$ were given values from $\{0.1, 0.2, $ $0.3, 0.4, 0.5, 0.6, 0.7, 0.8, 0.9\}$.\\\\
The methods were first tested on a simulation where the average number of trajectories traversing a link was, at any given time, around 30. This average was then increased by 100 and then by 200.\\\\
Table 2 shows the performance of the different methods in forecasting short-term traffic speed. Acronyms are used to describe the various exponential smoothing approaches; for instance, NSNT stands for no seasonality and no trend. We use root-mean-square error (RMSE) to evaluate performance, RMSE is defined as $$RMSE = \sqrt{\frac{\sum_{t=1}^n (\hat{y_t} - y_t)^2}{n}}$$
\begin{center}
\begin{table}[t]
{\small
\begin{tabular}{|p{2.63cm}|p{1.4cm}|p{1.45cm}|p{1.45cm}|}
\hline
&\textbf{ave = 30}&\textbf{ave = 130}&\textbf{ave = 230}\\
\hline
\hline
NSNT&5.135&5.080&4.869\\
\hline
NSAT&5.153&5.099&4.887\\
\hline
NSMT&5.172&5.109&4.903\\
\hline
ASNT&\textbf{4.752}&\textbf{4.761}&\textbf{4.577}\\
\hline
ASAT&5.029&5.054&4.805\\
\hline
ASMT&5.962&5.910&5.807\\
\hline
MSNT&4.849&4.857&4.678\\
\hline
MSAT&5.115&5.133&4.900\\
\hline
MSMT&5.051&5.077&4.863\\
\hline
Neural Network&\textbf{4.725}&\textbf{4.829}&-\\
\hline
Multiple Regression&\textbf{4.557}&\textbf{4.530}&\textbf{4.375}\\
\hline
\end{tabular}}
\vspace{3pt}
\caption{The performance of the different models based on RMSE. The top three models are highlighted.}
\label{tb:tablename}
\end{table}
\end{center}
\begin{figure*}
  \includegraphics[width=\textwidth,height=10.75cm]{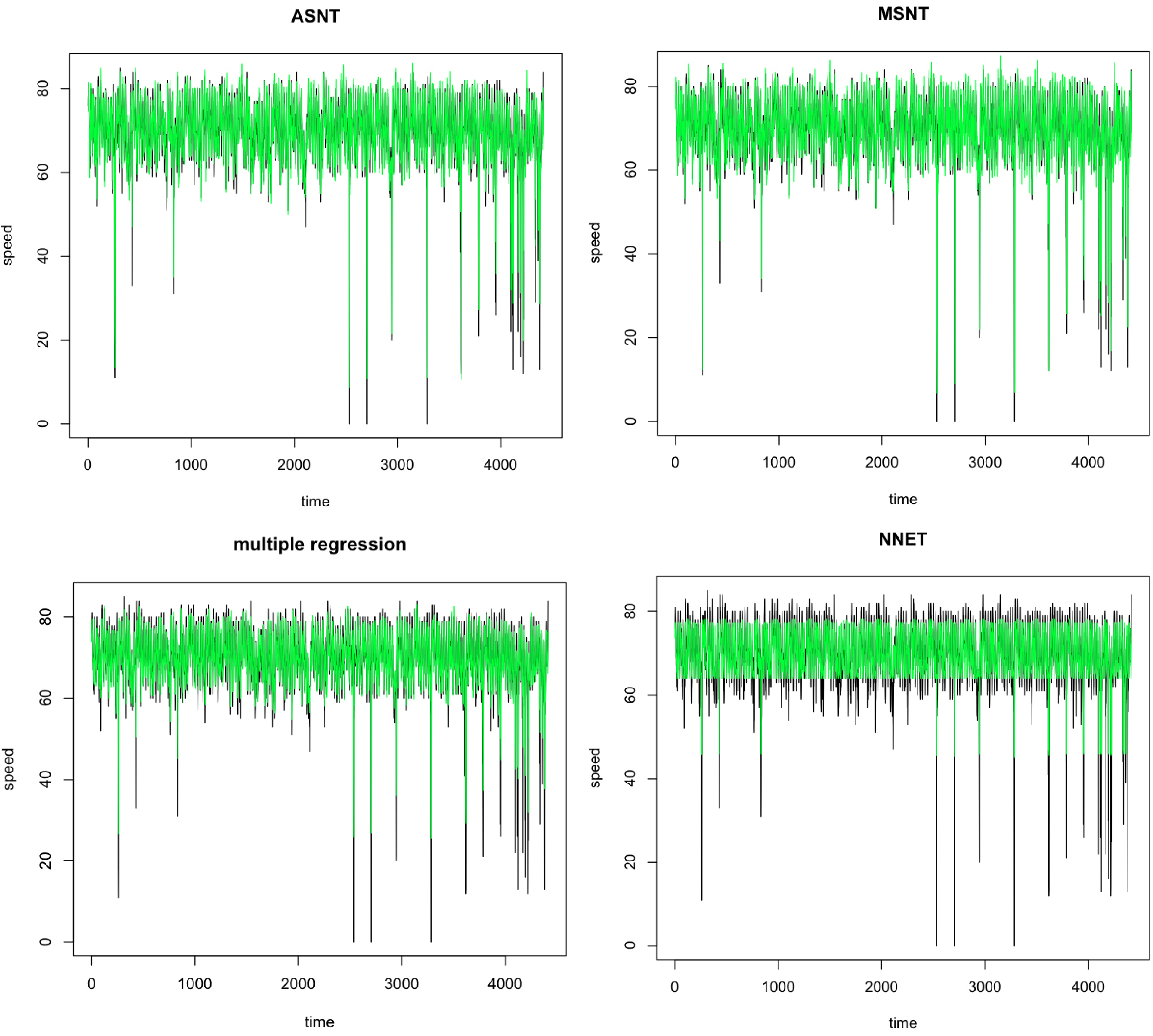}
  \caption{Single link traffic speed forecast using ASNT, MSNT, multiple regression, and neural network. }
\end{figure*}
where $\hat{y_t}$ and $y_t$ are the predicted speed and the actual speed at time $t$, respectively. RMSE is a good measure that tells us how far the predicted values are from the actual observed values.\\\\
As expected, the supervised learning approaches performed better in the evaluation. However, it is interesting to see that the performance of the methods based on exponential smoothing did not differ too much from that of the supervised techniques. In particular, the models with seasonality and no trend (additive seasonality and no trend, multiplicative seasonality and no trend) performed comparatively well and are way better than some of the other variants. This may be due to fact that even though traffic data does exhibit long-term trend, at the micro-level the traffic speed fluctuates frequently and does not show prolonged trend in one direction. Thus adding trend in the formula could be counterproductive in this case. In fact, when we simply use exponential smoothing without seasonality, the formulation that works best is the one without trend as well.\\\\
Figure 2 shows the forecasted values overlapped with the actual traffic speed values for the four best performing methods. It is interesting to note that the multiple regression seems to be able to predict traffic speed fairly accurately during typical conditions but almost always underestimates the speed when there are sharp drops or spikes in speed (outlier conditions). Adding additional predictors that indicate special events may help the model, however the question is how we can do this automatically. The exponential smoothing-based methods, on the other hand, seem to be more prone to overestimation of speed as evidenced by their graphs. Because of this, however, they are able to predict the speed fairly accurately during atypical conditions. The neural network forecasts, in this instance, seem to be the most conservative and are mostly centered around the mean. This may be due to the fact that neural networks are known to require large amounts of training data to perform well.\\\\  
Even though supervised techniques work well in forecasting, there is an overhead cost of retraining and discovering the ideal parameters of these models for every link in the road network. Given the potentially large amount of trajectory data that a real-world traffic forecasting system gathers continuously, it would be advantageous to use simpler models like ASNT or MSNT exponential smoothing models for forecasting at the cost of a slight reduction in accuracy.
\begin{center}
\begin{table}[t]
{\small
\begin{tabular}{|p{2cm}|p{5.7cm}|}
\hline
&\textbf{ideal parametric values}\\
\hline
\hline
NSNT&$\alpha = 0.9$\\
\hline
NSAT&$\alpha = 0.8, \beta = 0.1$\\
\hline
NSMT&$\alpha = 0.8, \beta = 0.1$\\
\hline
ASNT&$\alpha=0.5, \gamma = 0.1$\\
\hline
ASAT&$\alpha = 0.6, \beta = 0.1, \gamma = 0.1$\\
\hline
ASMT&$\alpha = 0.3, \beta = 0.1, \gamma = 0.2$\\
\hline
MSNT&$\alpha = 0.5, \gamma = 0.1$\\
\hline
MSAT&$\alpha = 0.6, \beta = 0.1, \gamma = 0.1$\\
\hline
MSMT&$\alpha = 0.7, \beta = 0.8, \gamma = 0.1$\\
\hline
\end{tabular}}
\caption{The optimal values for the parameters $\alpha, \beta,$ and $\gamma$.}
\label{tb:tablename}
\end{table}
\end{center}
\subsection{Ideal parametric values}
Table 3 shows the optimal values of the parameters used in the exponential smoothing models. For the models without seasonality, higher values for $\alpha$ are preferred which means that recently observed speed is more important in the forecast than long term information. On the other hand, lower values for $\beta$ are preferred which means that if trend is to be considered, it is best to update the trend gradually rather than update it immediately based on the slope of the last pair of observed speed.\\\\
It is interesting to note that in the model with additive seasonality and no trend, both the previous forecasted speed as well as the forecasted speed from the previous season are equally important. However, lower values for $\gamma$ and $\beta$ are preferred.\\\\
In fact, for all variants of exponential smoothing tested, low values for $\gamma$ yield the optimal results.
\section{Conclusion and some suggestions for future work}
In this work, we studied the problem of short-term traffic speed forecasting using vehicle trajectory data. Various models were evaluated and it was shown that learning based techniques outperform those based on exponential smoothing although certain variants of the second group of methods also performed quite well. The optimal values for the parameters of the various models were also identified and analysed.\\\\
We now list down some possible directions for future work. Since we were unable to collect a substantial amount of vehicle trajectory data for forecasting, we had to resort to using simulated data. In the future, we would like to test the models on real-world trajectory data. At the moment, we assume that the paths the trajectories take are all defined. In the real-world, however, most collected trajectories are uncertain which means that the actual path connecting two consecutive GPS coordinates is uncertain or unknown \cite{uncertain}. A traffic forecasting system will have to take this into account.\\\\
We would also like to consider testing other techniques for forecasting. In particular, we are interested in testing ensemble methods \cite{ensemble} which can yield better performance than stand-alone techniques when applied appropriately. Eventually, we plan to install systems on taxis as well as various public transportation vehicles which will collect the GPS trajectories. This will result in a large stream of data being uploaded continuously to the system's server, especially during peak hours. There is a need to develop a system which can handle the load. Furthermore, the system will ideally employ an online learning strategy \cite{online} so it can adapt dynamically.\\\\
Recent research have shown that spatial features are also important when forecasting short-term traffic condition \cite{pred1}. In future work, we would like to test the effect of incorporating spatially relevant features as input to our learning models.\\\\
Finally, it would be interesting to see how we can extend the models based on exponential smoothing to make them better. This can be done by combining the models with other techniques or by redefining or reformulating certain parts.
\section*{Acknowledgment}
This research was supported by a Commision on Higher Education-Philippine Higher Education Research Network (CHED-PHERNet) grant.

\bibliographystyle{abbrv}
\bibliography{sigproc}

\end{document}